Title:

# Agentic Language-to-Objective Synthesis for Optofluidic Assembly

Authors:

Ivan Saraev[1], Elena Erben[1], Weida Liao[2,3,+], Fan Nan[1], Gerhard Neumann[4], Eric Lauga[2], and Moritz Kreysing[1,*]

**1) Institute of Biological and Chemical Systems, Karlsruhe Institute of Technology, Germany**

**2) Department of Applied Mathematics and Theoretical Physics, University of Cambridge, UK**

**3) Department of Mathematics, Imperial College London, UK**

**4) Institute of Anthropomatics and Robotics (IAR), Karlsruhe Institute of Technology, Germany**

***e-mail: moritz.kreysing@kit.edu**

**Abstract**

Light-based advanced manufacturing increasingly requires programmable, closed-loop tools that translate human design intent into executable operations at small length scales. Yet a key bottleneck persists across robotic and manufacturing modalities: turning user intent into machine-readable objectives that are reliably executable. While micro-robotics offers versatile manipulation via optical actuation of fluids, mathematically tractable goal specification remains manual and hard to reuse. Here, we introduce Speak-to-Objective, a modular agentic pipeline that uses a conditioned Large Language Model (LLM) to translate spoken or written commands into fully differentiable objective functions for assembling microparticles in a constraint-aware inverse solver (SLSQP) and on an experimental optofluidic platform. The approach employs a compact loop—perceive → compose → propose → act → report & learn—that treats the objective as the interface between intent and actuation, separating what to assemble or pattern from how to actuate, while learning from user feedback. The pipeline composes geometry, spacing, and assignment/topology terms to generate robust descriptive objectives that assemble from partial traces and recover after perturbations, as well as explicit objectives for precise placement, all in an actuator-agnostic fashion. Using laser-induced thermoviscous flows as the physical actuation modality, we demonstrate natural-language-programmable, light-based microscale assembly of particle patterns in a microfluidic environment. Beyond its immediate impact on programmable microassembly, and using laser-induced optofluidic actuation as a reduced-complexity experimental platform, our work points toward self-driving, AI-assisted optical manufacturing platforms in which natural language, differentiable objectives, and laser-based actuation are coupled into a reusable digital workflow.

## Introduction

Microscale manipulation has progressed along multiple actuation paradigms, including optical tweezers[1-7], magnetic microrobots[8, 9], optically actuated micro-steerers[10, 11], micro-rockets[12], micro-tools[13], as well as Marangoni streamers[14, 15] and Janus-particle strategies[16], with each offering distinct trade-offs in force, speed, biocompatibility, and controllability[17]. Recent optofluidic approaches[18], which we explored in our previous work, complement these modalities by harnessing laser-induced thermoviscous flows[19-21] to position many particles in parallel with minimal material constraints, enabling rich, predictable flow topologies, even in highly viscous environments[22-25]. In the context of light-based advanced manufacturing, such optofluidic flows provide a reconfigurable optical actuation mechanism for contactless microscale assembly, patterning, and potentially biofabrication in liquid environments.

We previously demonstrated opto-fluidically multiplexed assembly using iterative, laser-induced flow fields choreographed via heuristics, even able to mimic high definition humanoid robotics[25]. We then moved to objective-based control via an analytical thermoviscous model. This study achieved sub-micrometer alignment and an emergent “action-at-a-distance” strategy that mitigates unwanted hydrodynamic coupling between particles, establishing that well-posed objectives combined with physics-aware optimization can replace rigid sets of rules[26].

Despite these advances in experimental automation, a key bottleneck remains: translating human intent (e.g. “arrange particles like a star”) into an executable objective function that a solver can minimize. Authoring such cost functions is time-consuming, brittle to task variation, and traditionally restricted to a specific actuation scheme, with a maximum of 15 particles being aligned in experiments or simulations[25].

Here, we seek to close the gap between human design intent and executable light-based microassembly by employing large language models to write the objective functions that bring particles into desired spatial configurations.. We introduce Speak-to-Objective, an end-to-end framework that compiles spoken or textual commands into differentiable cost functions. Conceptually, this provides a direct path “from thought to action” in which the objective function decreases the delay between intention and control. Concretely, we cast the front end as an agentic workflow that we summarize as “perceive → compose context → synthesize objective → execute → report & learn”. In it, the system greets the user (text-to-speech), transcribes intent with Automatic Speech Recognition (ASR), splices a user-rated example catalog with the new request and a system message, and inserts the response, an LLM-written cost function, into the program’s objective slot for execution. Because feedback after each run updates the few-shot catalogue, the workflow exhibits a form of lifelong learning; these catalogues can be recycled across model upgrades, so newer LLMs inherit effective strategies without changing the execution stack. In practice, the LLM agent emits compact, interpretable programs that combine external-shape, even-spacing, and topology/particle assignment terms; depending on the prompt, it may derive shapes analytically (e.g., pentagram geometry) or decide to invoke installed tools for vector outlines (e.g., a python library for extracting fonts such as matplotlib.textpath.TextPath) within a controlled execution context.

Similar pressures of labor-intense experimental planning have driven autonomous, ‘self-driving’ lab platforms across chemistry, physics and materials[27-31], including approaches that leverage LLMs for planning and control[32-33]. For example, *Eureka*[34] has utilized LLMs for reward function design in reinforcement learning. Here, the LLM is used in an interactive learning loop and proposes a population of several reward functions. Each reward is evaluated by running reinforcement learning on a given simulation environment and assessing the task-success of the final policy. This information is fed back into the LLM, in order to propose the next generation of reward functions that can then be used for reinforcement learning. Hence, the knowledge base of Eureka is also interactively extended; yet, the knowledge base is typically built up for the same task in an incremental manner.

*Eureka* has been extended to perform successful simulation-to-reality transfer in *Dr. Eureka*[35], where the sim-2-real difference for learned policies is also fed back to the LLM. Moreover, reinforcement learning[36], has been shown to be compatible with optofluidic navigation when trained on real experiments without explicit sensing.

Concurrently, the robotics community has begun employing foundation models at the language–vision–control interface. For example, recent work on scaffolding dexterous manipulation with vision–language models has used Vision Language Models (VLMs) to infer task-relevant key points and synthesize coarse 3D motion trajectories from natural-language descriptions and scenes, which are then refined by low-level controllers[37].

Our approach is complementary: rather than proposing reference trajectories or policies, we use LLMs to synthesize the objective itself, yielding an interpretable, actuator-agnostic interface between intent and physical execution that makes the objective come true. Within this agentic loop, user feedback after each run updates the example catalogue, improving future objective synthesis without altering the execution stack. This separation clarifies roles: language specifies success criteria, while the solver determines how to actuate under the chosen physical constraints.

On the execution side, we pose the inverse problem over a multi-flow superposition model and solve it using Sequential Least Squares Programming (SLSQP) under bounds and constraints. This replaces our earlier random search approach[26], while improving convergence speed, constraint compliance (e.g.minimum spacing), and scalability to complex shapes. Execution proceeds along two pathways: (i) a generic potential–based solver for catalogue evaluation and complexity sweeps (Fig. 2), and (ii) a constraint-aware inverse solver over thermoviscous flow superposition with LUT acceleration and SLSQP (Fig. 3); the identical objective can also drive the experimental platform (Fig. 4), enabling direct simulation–experiment comparisons. Beyond analytic (explicit) objectives, we leverage descriptive objective functions rather than explicit coordinates that confer self-healing after perturbations and rapid assembly from partial traces, while retaining differentiability and auditability. Together, these elements establish an agentic, interpretable, and reusable path from natural-language intent to physical assembly. From an advanced-manufacturing perspective, the key contribution is not a new particle material, but a programmable control layer for optical microassembly: natural-language design intent is converted into differentiable objectives, which are then realized through laser-induced flow fields. This positions the platform as a digital production tool for reconfigurable microscale patterning, particle organization, and future light-assisted biofabrication workflows.

**System Overview — Agentic workflow from intent to execution**

Our pipeline operates as an effective tool-using agent that converts a natural-language request into an objective functional and executes it via a chosen forward model. The cycle is: perceive → compose context → synthesize objective → execute → report & learn.

The agent greets the user via text-to-speech, explains its capabilities, and solicits a task; automatic speech recognition captures the wish of the user (e.g., "Arrange particles like a star"). It then combines a system message (LLM is told: "You are an agent to write objective functions; here are examples and the new task….") by splicing with a compact, user-rated example catalogue—which already includes user feedback from previous rounds—with the new request ("Arrange particles like a star").

Conditioned on this context, a language model synthesizes a compact cost function f(A), where A is an array of n particle positions in 2 dimensions. The program extracts the code block containing f(A) and inserts it into the objective slot of the pipeline. An example of such a two-dimensional cost function can be found in Figure 1, where a star is specified by geometric principles.

Execution proceeds along one of two pathways, selected by the user or the experiment script as detailed further below: (i) A generic numerical potential solver (used for the catalogue tiles and complexity analysis in (Fig. 2) minimizes f(A) to produce steady-state particle arrangements from randomized initial conditions in a permutation-invariant manner. (ii) A constraint-aware inverse solver over thermoviscous flow superposition (Fig. 3) minimizes the same f(A) while optimizing multi-flow parameters (LUT-based, SLSQP), i.e. collective particle motion is constrained via the velocity fields specified within the LUTs. In the experimental branch (Fig. 4), the identical objective drives the optofluidic setup (described before in [26]) and leads to real-world microparticle assembly.KI

After each run, the agent collects user feedback and these ratings flow back into the example catalogue, improving future synthesis without altering the execution stack. As a result of this, in addition to functional example (DOs) the catalogue also contains failed attempts (the DON'Ts).

Treating the objective function as the interface preserves interpretability, i.e. for scientific purposes, enables actuator-agnostic reuse (swap the forward operator; keep the function), and closes an iterative loop in which user feedback continually refines the agent's proposals.

We evaluated the agentic synthesis–execution loop on a catalogue of language prompts spanning primitives (lines, circles), polygons, stars, and composite motifs. The LLM is conditioned on a user-rated few-shot catalogue and proposes an objective function f, which is executed by the generic potential–based solver from randomized initializations.

Performance is measured by a vision model (Open AI O3) that is asked whether the final arrangement matches the user's pattern description: the scorer returns a similarity/confidence S in interval [0,1]; the user rates the outcome by human language (e.g. "the star looks good but is very small"). With a catalogue of >10 rated examples, first-shot success typically exceeds 50% (mean over tasks), with simple shapes (circles, lines) approaching a ceiling and more complex shapes degrading gradually rather than abruptly.

Increasing the few-shot context from 0→5→10→20 examples yields fast gains; ironically, the very steep learning curve makes precise quantification difficult, and once the catalogue passes ~10 examples, success rises to very high levels for simple objectives. *Figure 2 (top)* shows vision-rated success versus complexity, annotated with representative outputs as the horizontal axis progresses from “a dashed line” through “an ellipse” to “a dashed sinusoidal line” and “squares arranged like the points of a hexagon” with insets showing solver outputs for intermediate cases. Success rate ranges between >85% success for low complexity tasks to about 40% complex hierarchical arrangements. Example outputs and their variability are shown in Figure 2b.

**Synthesized objective functions: design patterns returned by the agent**

The language model returns a compact objective as fenced Python that is cut out and inserted into the program’s objective slot. In practice, we find that each objective is a weighted sum of normalized terms drawn from the three guiding concepts described below, covering most prompts, from “circle” and “pentagram” to semantically richer requests:

A) **External shape: where particles should go.**

- **Coordinate-specific definition.** Move particles toward sampled points along the outlines of letters and polygons; distances are computed to masks/skeletons and normalized for scale/rotation when needed. Interestingly, the pipeline can use different strategies for writing that code, and invokes external libraries that bring in specific tools, such as the rendering of font via the matplot library.
- **Shape-descriptive definition.** For relation-defined shapes (e.g., a square), enforce equal side lengths, right angles, and similar constraints; optional soft anchors (center/scale/orientation) may pin shapes to positions on a grid.
- **Parametric curves.** Use closed-form primitives (circles/ellipses, sinusoids, spirals, hearts) sampled by arc length; minimize distance of particles to the curve or its skeleton.

B) **Even spacing: make it look tidy.**

- **Short-range repulsion.** A soft barrier between nearest neighbors promotes uniform spacing (Poisson-disk-like packing).
- **Discrete/topological sites.** When a lattice or perimeter is intended, one-per-site assignments (e.g., equally spaced points on a ring or along polygon edges) make spacing emerge automatically.

C) **Assignment: who goes wher**e

- **Region vs. periphery.** Bias a chosen subset to an interior mask while steering the remaining particles to the boundary (or onto a ring with large radius).
- **Assignment / optimal transport.** In case of coordinate-specific objective functions, particles are assigned to nearest targets.

This triad contributes to clean, interpretable assemblies and can be used for the generic solver or the thermoviscous multi-flow solver. An example of the assembly can be seen Supplementary Movie 1.

### Alternative strategies to arrive at right geometric shapes

When asking how the language model "finds" the right shapes, we observe this to be dominated by three complementary strategies, spanning hand-crafted / intuition-based geometry, analytic calculation, and font-based rendering, or combinations of these.

As a concrete instance of the geometry/analytic pathway, the agent was prompted to produce "three hexagons that touch but don't overlap". It set a side length s, placed three centers at offsets [0, 0], [1.5s, $\sqrt{3}$ / 2 s], and [3s, 0] to ensure edge-to-edge contact, then constructed six vertices around each center at 60° increments. The perimeter of each hexagon was sampled by evenly distributing particles along its edges, with particle counts balanced across all 18 edges and excess handled by round-robin assignment. The three sets of boundary points were combined, recentered around the origin, and taken as the target lattice. The resulting deviation between actual particle coordinates and these idealized edge points defined the cost, enforcing a tidy arrangement of particles along the triple-hexagon motif.

Independently, simple lettering (e.g. "**K I T"**) was sometimes constructed from line-by-line defined segments to generate stroke templates, indicative of general intelligence and geometric understanding (Supplementary Movie 2.).

For more complex patterns, like letters or glyphs, the LLM agent frequently followed a tool-assisted rendering route. It thereby invoked installed libraries such as matplotlib.textpath, which it started to invoke by itself throughout the learning process.

Together, these pathways show that the system utilizes thinking capabilities about geometry when that is most efficient, or employs tools when these offer an advantage, and then use either result in the same objective-synthesis framework (shape + spacing + assignment).

### Inverse problem flow solver

In our flow simulations, repeated scanning of a laser beam induces inertialess, thermoviscous fluid flow and hence net transport of particles[24]; to good approximation, the velocity fields resulting from sequential scanning of multiple paths add linearly[25]. We exploit this by precomputing the flow response to the canonical linear scan path (typically 10um long) into a lookup table (LUT) and reading velocities by fast interpolation during optimization.

### From random search to constrained continuous optimization

In comparison to the random-search controller used previously[26].—which draws a single scan path (center + orientation) and accepts it only if it lowers the objective by a threshold amount—our framework replaces proposal-and-accept with continuous, and parallel multi-scan-path optimization. We use scipy.optimize.minimize (SLSQP) with the LUT flow model, seeking N flow fields to be applied simultaneously (per flow field: coordinates for center of scan path, angle of scan path, optionally amplitude) under explicit constraints and bounds (minimum spacing between sub-flow field centers, keep-out regions). During optimization, velocities from all sub-flows are superimposed to produce coordinated actuation. As that, we optimize a N*3 dimensional multi-constraint problem, thereby accessing previously described experimental capabilities for multiplexing[25].

The practical consequences of this advancement in optimization are multifold. The new formulation yields: (i) greater control per feedback step (topologically rich flow fields instead of sequential single paths); (ii) fewer model evaluations for the same decrease in objective (SLSQP follows a descent direction rather than blind resampling); (iii) finer control via additional constraints (field of view, minimum inter-particle spacing, exposure limits); (iv) deterministic, warm-startable behavior (or informed guess where to start searching) for reliable convergence; and (v) perspectively a clear path for parallelization, since the use of a LUT and vectorization map well to GPUs and automatic differentiation (although not needed so far).

Moreover, we found the optimization process to be actuation-pattern-agnostic. This means that while we originally conceived the flow pattern solver to work downstream of the objective function synthesis using linear scan paths, we found that optimization also still converges when choosing other flow fields as actuation primitives, such as circular scan paths[39] (fig 3c), saddle-point type flow fields (fig 3d) or flow field configurations with increased local shear and richer topology (fig 3e).

In summary, by treating the objective function as a stable, interpretable interface between human design intent and optical actuation, we decouple “what to assemble” from “how to actuate”, thereby closing the gap between language-level specification and light-based microscale assembly.

## Experimental assembly and *phoenix-from-ashes*-type behavior

We next tested LLM-generated objectives in experiment, using the same optofluidic platform and imaging hardware as in our previous work[26]. We prompted the agent for a square-shaped assembly. Rather than producing static target coordinates, the model returned a descriptive (shape-specific) objective function that enforces the geometric constraints of a square (equal side lengths and right angles), while leaving point-specific parameters (position, orientation, and—under normalization—scale) free. Because the score measures relative geometries rather than absolute positions, it is naturally invariant to translation, rotation and even size scaling, while remaining tractable by gradient-based solvers. In practice, the optimizer searches for any square consistent with the objective, rather than for one fixed at predetermined coordinates.

This design has four experimentally useful consequences over previous strategies we used. (i) Minimal, local motion. From arbitrary initial distributions, particles are softly matched to nearby sites on the square, so they move as little as necessary—they “glide” relative to each other rather than traverse long detours. (ii) Automatic localization within the field of view of the microscope. Even without an explicit centering term, the solution by definition localizes inside the camera window (avoiding size scaling problems). (iii) Invariance under rigid/scale transforms. Any translated/rotated (and, with normalization, resized) square is scored as correct. (iv)

After a transient perturbation that displaced a subset of particles (or even transformed the square into a triangle), the assembly re-converged to a square with slightly different size and orientation , in a *phoenix-from-ashes* type manner (Figure 4b): the objective value rose at the disturbance and then decayed; a squareness index recovered to baseline within a few control cycles. In an alternative realization, the program generated the square with about half the area, yet still satisfied the descriptive criterion. By contrast, an explicit, coordinate-anchored objective is more more brittle and inefficient in restoring the original shape: it would try recover the original absolute positions even when squareness is already restored (e.g. after uniform translation).

For many assembly tasks, relations (equal edges, right angles, inside/outside) matter more than absolute coordinates; the descriptive formulation aligns with that need. It also accelerates completion from partial traces: when initialized with only fragments of a square, the pattern closes quickly because the objective supplies a shape prior rather than a fixed map of destinations.

A similar example of this type is the accumulation of particles, i.e. in the center of the field of view to locally reach higher concentrations that exceed both the mean as well as spontaneous fluctuations from it. Here, especially for interchangeable particles, the ensemble state matters more that the precise individual positioning. Towards this end we find that the LLM planned experiments readily succeeds in locally increasing the concentration of particles (fig 5), well beyond chance. Specifically, the locally generated assembly goes together with an about 15 fold increase in particle density, a state that would spontaneously be observed with a probability smaller than one in ten billion.

## Discussion:

### From design intent to light-based microassembly: the objective as bridge between language and execution

This work addresses the language-to-manufacturing-control gap by letting an agentic LLM translate a natural-language design request into an objective function that a solver can optimize and execute through a chosen optical forward model or experimental light-based actuation platform. Treating the objective as a module decouples “what to achieve" from “how to actuate”: users specify criteria for success (shape, spacing, topology), while the system selects controls consistent with those criteria. This mirrors actuator-agnostic abstractions in autonomous experimentation and closed-loop optimization[38]. In our case the advantages are threefold. First, interpretability: terms and weights (pre-factors) can be inspected, altered or re-weighted like any other model component. Second, purpose and actuator-agnostic reuse: the same specification runs with a generic potential solver, a thermoviscous flow model, or the physical setup by swapping only the forward operator, while simulations additionally show the particle assembly task to be flow pattern agnostic. Third, comparability and auditability: because the objective is explicit, simulation–experiment comparisons become direct. In practice, the agentic loop—perceive → compose → propose → act → report & learn—proved effective: user ratings update the catalog and improve first-shot success without touching the execution stack (Fig. 1). This creates a form of lifelong learning in which accumulated examples and strategies persist beyond a single session and can be recycled: newer LLMs can be conditioned with the same catalogues as a starting point to inherit useful patterns.

We consistently observe compact programs that combine external shape, even spacing, and topology/assignment terms. These objectives are differentiable and stable across simulation and experiment (within the framework of current experimental capabilities), yet they also exhibit geometric intuition: for some prompts, the agent derives shapes analytically (e.g., a cartoon pig), while for text and complex glyphs, it invokes installed tools (e.g., matplotlib.textpath.TextPath) to extract vector outlines. The descriptive (pose-agnostic) variant is particularly powerful for assembly: it confers translation/rotation/scale invariance, is by definition within the field of view, and self-heals after perturbations; objectives using absolute coordinates, in turn, offer tighter positional control when it is required.

While our approach shares algorithmically similar components, the type of cost functions, systems and tasks considered in Eureka[34,35] are very different, as our objective functions are directly executable, frequently in their very first version, whereas a large set of reward functions written by Eureka are the basis for training a model.

## Strengths, limitations, and outlook

Replacing single-path random search[26] with SLSQP over a LUT-based flow model yields fewer evaluations and shorter runtime, with benefits relevant especially for more complex patterns. As a characteristic of this approach, physical interpretability is retained: optimized controls remain superpositions of simple scan elements (e.g., N=5 paths → a 15-parameter decision vector), and the observed net field is the direct sum of sub-flows. While not necessarily finding the global optimum, the solver is deterministic and can leverage warm-starts, or educated guesses (scan paths closest to outliers) for faster convergence. Core constraints, such as enforcing a minimal spacing between flow fields, are reflected centrally in the optimization process.

There remains, however, room for further developments.

A) Sim-2-real gap. As common in the field, experiments are clearly not yet at the height of the simulated results, especially regarding particle number, Further refinements on the experimental side will be required in the future.

B) Assembly strategies are clearly not always near the global optimum: especially for more complex structures, where it requires only one strong flow field to break the structure, solvers cannot always rebuild it in one step even though physics is deterministic and time-reversal symmetry would enable to this option. Here, better inverse solvers including multi-timestep planning will be useful, which could also prevent transient jamming states.

C) With an average first-shot success rate of >50%, there remains room for improvement, which will likely be enabled by future models of LLMs and increased context windows to accommodate larger catalogues.

D) Pipeline architecture: additional agents for quality control (i.e. a critical reviewer that requests targeted revisions when the vision score is low), safety (constraint auditor), and experiment orchestration (latency-aware scheduling, automatic parameter sweeps). This will become relevant, as experimental control improves and faster imaging and lower-latency actuation will be needed to keep pace.

Recent agentic AI systems have shown how language-model-based agents can support scientific discovery by generating hypotheses, analysing data, planning experiments, or writing expert-level empirical software[40–42]. Speak-to-Objective and the downstream inverse-optimization layer will integrate with such frameworks by providing an executable interface through which higher-level discovery agents can translate human- or agent-generated goals into optofluidic hardware actions. In this role, the objective function becomes a transparent hand-off layer between scientific reasoning, numerical optimization, and physical microscale motion control.

In summary, with this work, we close the language-to-control gap for light-based optofluidic assembly by using a simple LLM agentic pipeline that can invoke installed libraries to synthesize an objective from natural language and then execute it through laser-induced flow fields. Framing the objective function as an explicit module lets language specify “what success looks like” rather than “how to get there”, enabling interpretable, actuator-agnostic, light-based microscale assembly.

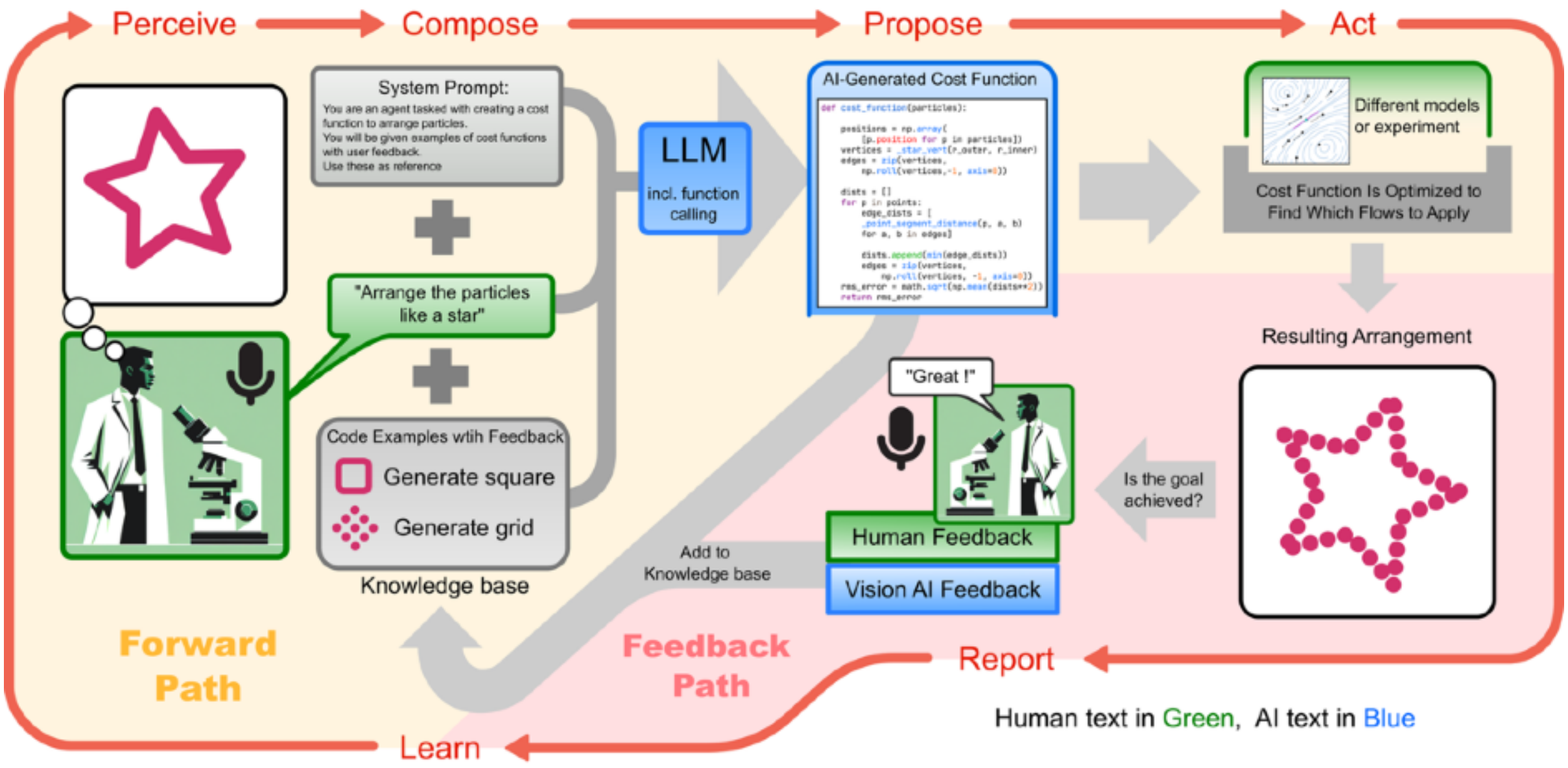

**Figure1: Agentic intent-to-light-based-assembly framework:** A user issues a natural-language request (green; speech or text), which is transcribed and combined with a system prompt and a few-shot knowledge base of prior objective functions together with ratings. The LLM generates an objective function that encodes the task in terms of shape/spacing/topology. This objective is inserted into the program's objective slot and optimized over a model of thermoviscous flows (in simulation or experiment) to determine which flows to apply, yielding the resulting arrangement as shown at the bottom right corner. After each run of the simulation, Vision-AI scoring and human feedback update the knowledge base, such that future prompts benefit from improved examples without changing the execution stack.

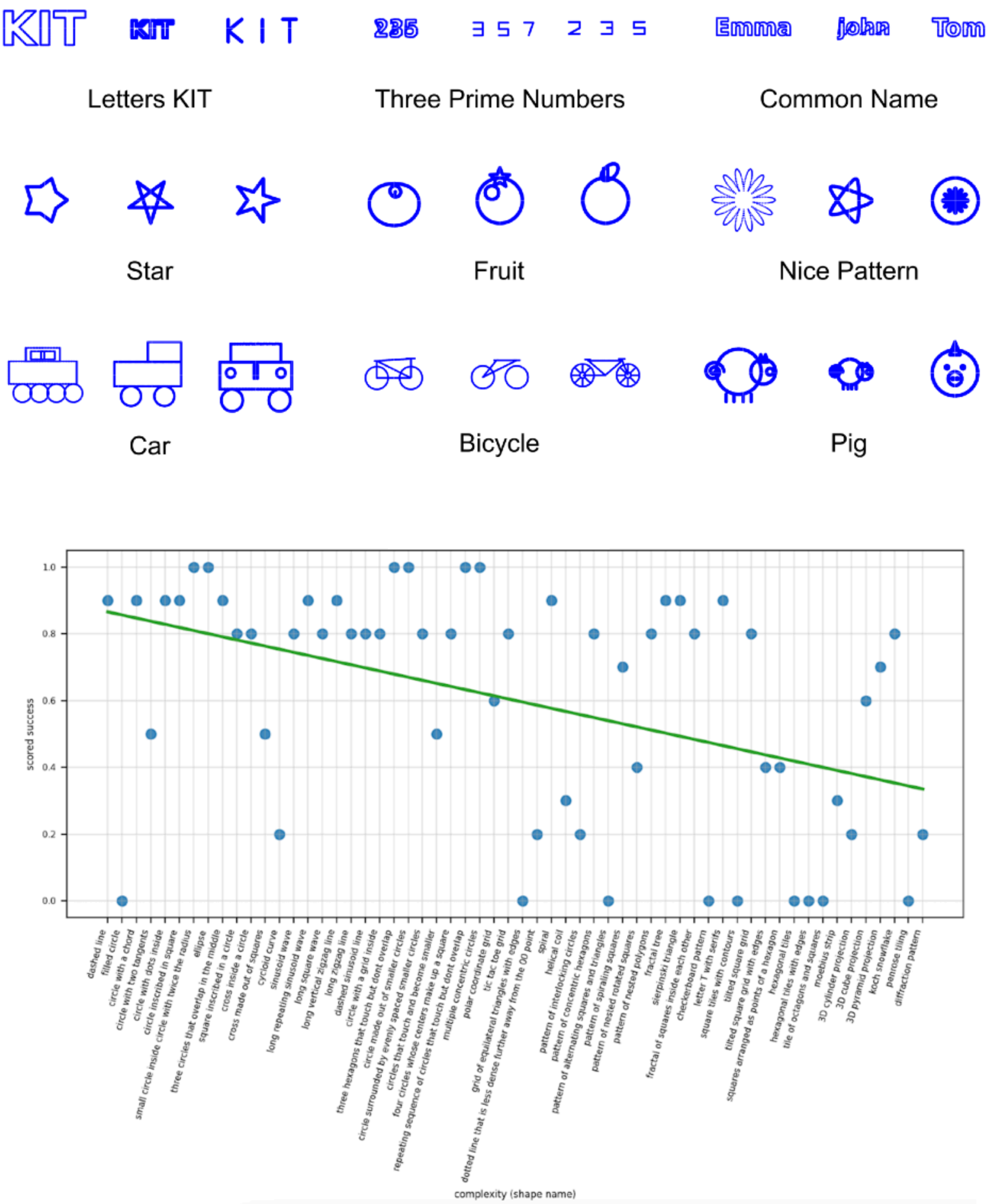

**Figure 2 | Catalogue and success rate across assembly task complexity. (a)** Examples of synthesized objective functions. Representative tiles from the knowledge base: each shows the natural-language prompt (excerpt, i.e. “the letters KIT”), and the resulting configuration from the generic potential solver, for 3 independent LLM calls. **(b)** Success rate vs complexity. Vision-model-rated success versus pattern complexity. Bars from left to right summarize primitives (triangle, square, circle) with high success; the green line tracks mean success as tasks become more complex (illustrated, for example, with a closed curve and a “pig with a spiral tail”).

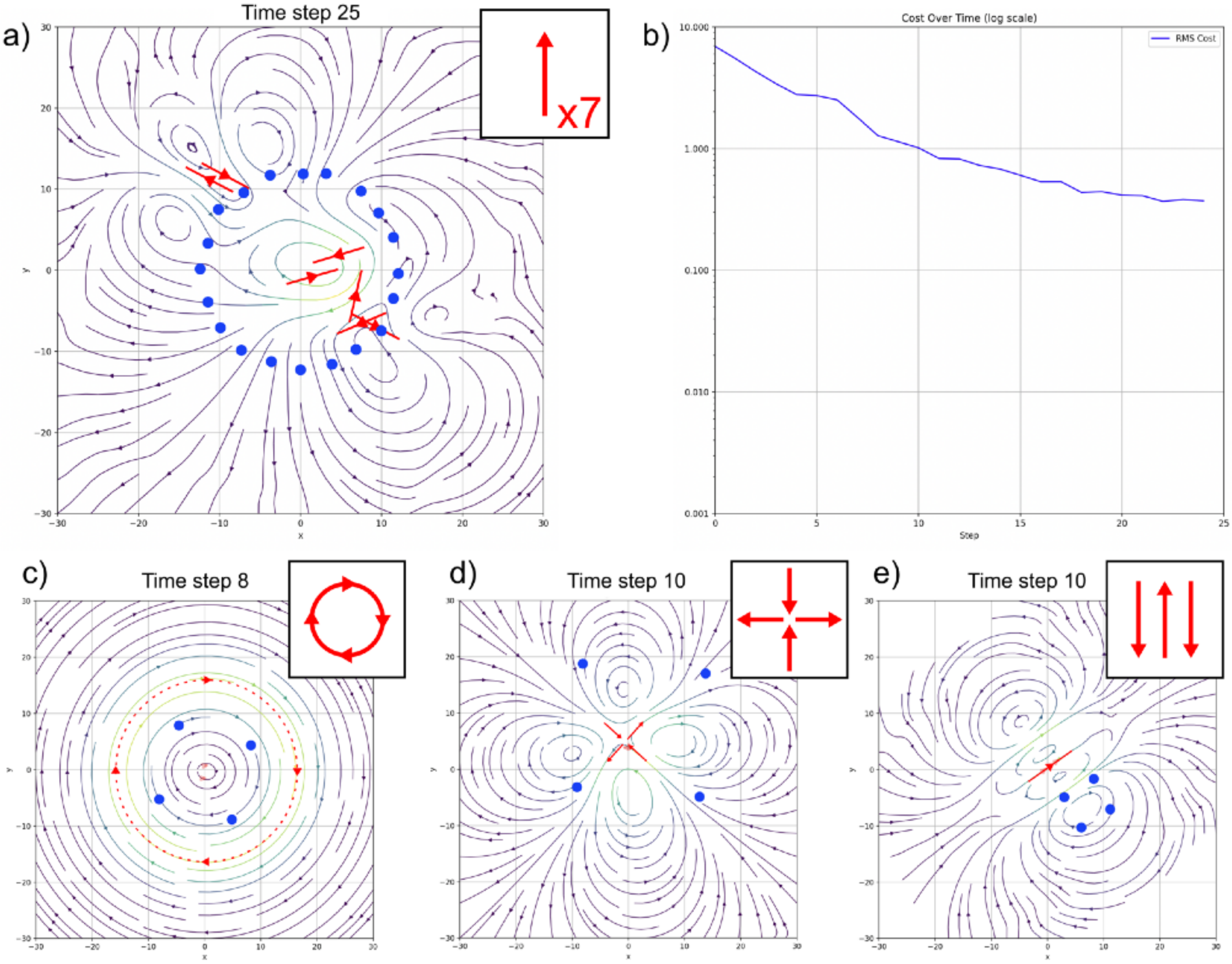

**Figure 3 | Inverse optimization for laser-induced microscale assembly:** Computing scan paths via inverse optimization. Flows move particles into alignment with the specified objective. Using the cost function generated by the language model, the optimizer autonomously selects scan paths. **(a)** 7 linear scan paths are selected and overlapped to arrange 20 particles as a circle. Here the optimization relies on SLSQP operating over a precomputed lookup-table (LUT) flow model, where each of the N = 7 scan paths contributes three parameters (position, orientation, amplitude), yielding a 21-dimensional optimization problem. The effective control field emerges as the superposition of the individual sub-flows, producing complex yet physically consistent actuation patterns from mathematically grounded primitives. **(b)** Cost metric of circle versus time in simulation steps. **(c-e)** Demonstration of actuation-agnostic alignment capabilities. Laser scan paths are shown in top right corners. **(c)** The actuation pattern is switched to a circular flow field to align four particles as a square after 8 steps of the simulation. **(d)** The actuation pattern is switched to a “saddle point” flow field and achieves alignment in 10 steps. **(e)** The actuation pattern is switched to a “shearing” flow field and achieves alignment, here at time step 10.

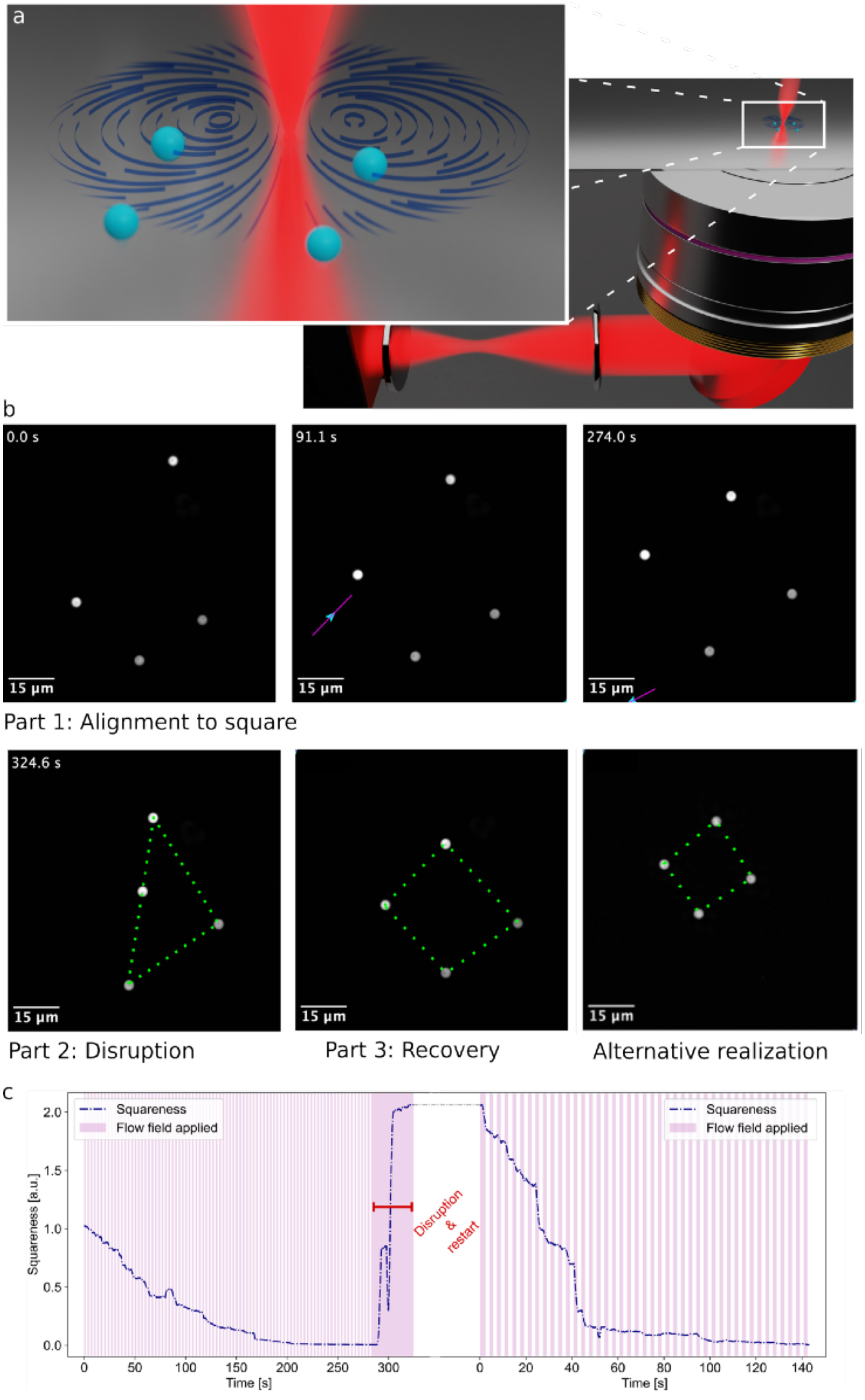

**Figure 4 | Experimental realization of light-based microscale assembly using a descriptive objective function (a)** Schematic of experimental setup. A mildly heating laser beam (red) is scanned through a fluid to induce thermoviscous flow (blue) that align particles (cyan) according to the objective function**. (b)** Time-lapse frames from a single experiment. *Part 1: Alignment to square*—starting from arbitrary positions, particles are steered by accepted scan paths (magenta vectors) toward a square; the fitted target skeleton is shown (green dashed). *Part 2: Disruption*—an external perturbation transiently deforms the pattern into a triangle (red overlay). *Part 3: Recovery and alternative realization*—with the **shape-descriptive objective**

unchanged (no explicit centering or pose terms), the system re-assembles a valid square, that may change size in alternative realizations. Scale bar: 15 μm; timestamps in seconds. **(c)** Squareness cost metric versus time. Shaded bands mark actuation cycles. **Left:** full trial showing initial assembly, a sharp loss of squareness at the perturbation, and autonomous recovery. **Right:** zoom on the recovery segment, highlighting near-monotonic improvement over a few control cycles. The final square may differ in orientation and size from the initial configuration, consistent with the objective's pose-agnostic design.

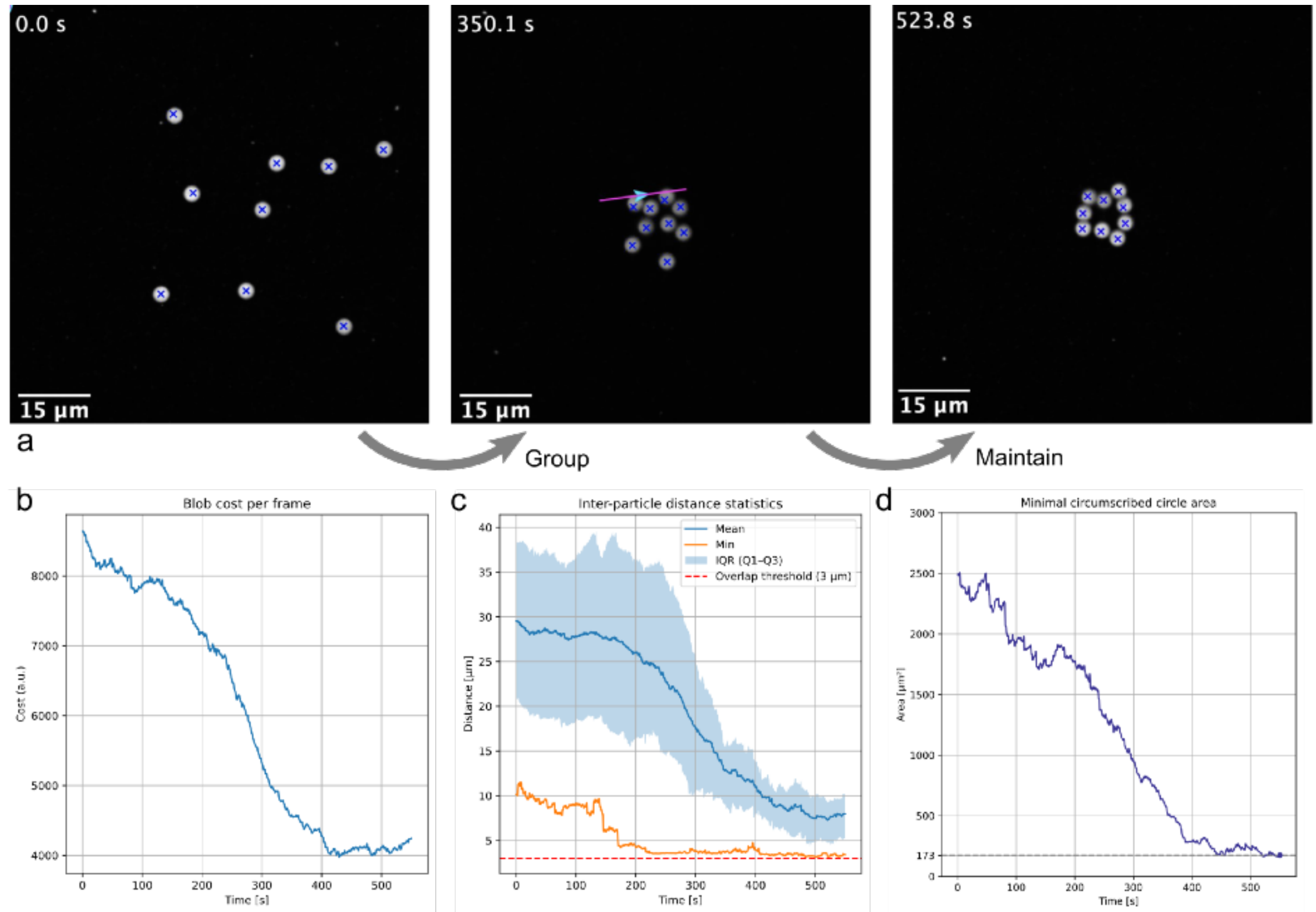

**Figure 5: Descriptive-objective control for natural-language-programmable optical concentration of microparticles. (a)** A natural-language request is translated by the agent into a **descriptive, ensemble-level objective** that increases particle density within a user-defined target region, without assigning absolute target positions to individual particles. **(b)** Starting from an initially dispersed distribution, laser-induced thermoviscous flows drive particles toward the target region in a closed-loop optofluidic assembly process; only the collective concentration state is scored, so particles remain interchangeable. **(c)** Time-lapse images showing the emergence of a localized particle accumulation under objective-driven optical actuation. **(d)** Quantification of local particle density relative to the field-of-view average and to random fluctuations. The optically assembled state reaches an approximately 15-fold density increase, corresponding to a configuration with a spontaneous occurrence probability below. This experiment illustrates that the same language-to-objective framework can encode not only geometric patterns, but also descriptive collective objectives in which ensemble properties, rather than absolute single-particle coordinates, define successful microscale assembly.

Caption Supplementary Movie 1:

Movie illustrating solution to Inverse problem: which sequence of flow fields is suitable to align 20 suspended particles as a circle.

Caption Supplementary Movie 2:

Letters “KIT” being assembled by a generic potential solver. 500 particles, aligning according to fully differentiable LLM written objective function.

**Acknowledgements:**

Acknowledgements: The authors acknowledge funding by the Karlsruhe Institute of Technology and the University of Cambridge. M.K. further acknowledges support by the European Research Council, in particular the ERC Starting Grant GHOSTs (Grant No. 853619), and by the Hector Foundation. M.K. also acknowledges support by the Volkswagen Foundation (Life! Grant No. 92772). The Kreysing lab was co-funded by the Deutsche Forschungsgemeinschaft (DFG, German Research Foundation) under Germany's Excellence Strategy—2082/1–390761711 (3DMM2O), Research Grant 515462906, and the DFG project 7593. The authors also acknowledge support by the Karlsruhe School of Optics & Photonics (KSOP) and the Helmholtz Program "Natural, Artificial and Cognitive Information Processing" (NACIP). W.L. gratefully acknowledges funding from the Engineering and Physical Sciences Research Council (EPSRC studentship) and Trinity College, Cambridge (Rouse Ball and Eddington Research Funds travel grant). The authors thank their colleagues for valuable feedback and discussions, and previous members of the research group for establishing the fundamental framework for the experimental setup and control software for thermoviscous flows. In this work, we used generative-AI tools in a limited and transparent manner. First, we used an API-based system (ChatGPT o1-preview and ChatGPT-4o) to assist in drafting and annotating cost functions, and as a coding assistant for autocomplete, code structuring, and commenting. Second, we used cursor.ai as an integrated coding assistant to help write, refactor, and debug the code pipeline according to our ideas. Third, after completion of the technical work, GPT-5 was used to improve the language and clarity of a manuscript conceived and written by the human authors (tone adjustment, proofreading, and style feedback). All conceptual design, experimental work, data collection, analysis (except where explicitly stated otherwise, e.g. automated scoring of complexity in Fig. 2b), and interpretation were carried out by the human authors, who critically reviewed and approved all content. Moritz Kreysing wishes to thank Dr. Steffen Grosser for feedback on the first steps in this research.

**Author contributions:**

Coding and implementation: IS, MK. Experiments and their analysis: EE, IS. Analytic actuation patterns and flow primitives: EL, WL. Conception of project: MK. First manuscript draft: MK, completed with expert input from GN (robotics and machine learning), FN (optical micromanipulation), and IS, and critically discussed by all authors.

**Data availability:**

All data supporting the findings of this study are generated by the code, which was submitted with the publication and is publicly available for download at **[persistent link or DOI, XXX upon acceptance]**; the microscopy images acquired and/or analysed during the current study are available from the corresponding author upon reasonable request.

**Conflict of interests:**

E.E. and M.K. previously applied for an international patent for optofluidic technology related to this publication (application number PCT/EP2021/071437). M.K. is a co-inventor of technology for laser induced flow and force measurement technology (US Patent App. 17/506,750 and PCT/EP2021/071392) and holds a consultancy contract with Rapp Optoelectronic GmbH

1. Ashkin, A.; Dziedzic, J. M.; Bjorkholm, J. E.; Chu, S., Observation of a single-beam gradient force optical trap for dielectric particles. Opt. Lett. 1986, 11 (5), 288-290.

2. Grier, D. G., A revolution in optical manipulation. Nature 2003, 424 (6950), 810-816.

3. Roichman, Y.; Sun, B.; Roichman, Y.; Amato-Grill, J.; Grier, D. G., Optical forces arising from phase gradients. Phys. Rev. Lett. 2008, 100 (1), 013602.

4. Chen, J.; Ng, J.; Lin, Z.; Chan, C. T., Optical pulling force. Nat. Photon. 2011, 5 (9), 531-534.

5. Maragò, O. M.; Jones, P. H.; Gucciardi, P. G.; Volpe, G.; Ferrari, A. C., Optical trapping and manipulation of nanostructures. Nat. Nanotechnol. 2013, 8 (11), 807-819.

6. Gao, D.; Ding, W.; Nieto-Vesperinas, M.; Ding, X.; Rahman, M.; Zhang, T.; Lim, C.; Qiu, C.-W., Optical manipulation from the microscale to the nanoscale: fundamentals, advances and prospects. Light Sci. Appl. 2017, 6 (9), e17039-e17039.

7. Li, X.; Dan, D.; Zavatski, S.; Gao, W.; Zhang, Q.; Zhou, Y.; Qian, J.; Yang, Y.; Yu, X.; Yan, S.; Xu, X.; Martin, O. J. F.; Yao, B., Optical tweeze-sectioning microscopy for 3D imaging and manipulation of suspended cells. Sci. Adv. 2025, 11 (27), eadx3900.

8. Zhou, H.; Mayorga-Martinez, C. C.; Pané, S.; Zhang, L.; Pumera, M., Magnetically Driven Micro and Nanorobots. Chemical Reviews 2021, 121 (8), 4999-5041.

9. Kim, S.; Qiu, F.; Kim, S.; Ghanbari, A.; Moon, C.; Zhang, L.; Nelson, B. J.; Choi, H., Fabrication and Characterization of Magnetic Microrobots for Three-Dimensional Cell Culture and Targeted Transportation. Advanced Materials 2013, 25 (41), 5863-5868.

10. Būtaitė, U. G.; Gibson, G. M.; Ho, Y.-L. D.; Taverne, M.; Taylor, J. M.; Phillips, D. B., Indirect optical trapping using light driven micro-rotors for reconfigurable hydrodynamic manipulation. Nat. Commun. 2019, 10 (1), 1215.

11. Phillips, D. B.; Padgett, M. J.; Hanna, S.; Ho, Y. L. D.; Carberry, D. M.; Miles, M. J.; Simpson, S. H., Shape-induced force fields in optical trapping. Nature Photonics 2014, 8 (5), 400-405.

12. Li, D.; Liu, C.; Yang, Y.; Wang, L.; Shen, Y., Micro-rocket robot with all-optic actuating and tracking in blood. Light: Science & Applications 2020, 9 (1), 84.

13. Zhang, S.; Xu, B.; Elsayed, M.; Nan, F.; Liang, W.; Valley, J. K.; Liu, L.; Huang, Q.; Wu, M. C.; Wheeler, A. R., Optoelectronic tweezers: a versatile toolbox for nano-/micro-manipulation. Chem. Soc. Rev. 2022, 51 (22), 9203-9242.

14. Dara, P.; Käll, M., Bubble Dynamics and Directional Marangoni Flow Induced by Laser Heating of Silicon Nanodisk Arrays. The Journal of Physical Chemistry C 2025, 129 (11), 5502-5510.

15. Dara, P.; Shanei, M.; Jones, S.; Käll, M., Directional Control of Transient Flows Generated by Thermoplasmonic Bubble Nucleation. The Journal of Physical Chemistry C 2023, 127 (35), 17454-17459.

16. Nedev, S.; Carretero-Palacios, S.; Kühler, P.; Lohmüller, T.; Urban, A. S.; Anderson, L. J. E.; Feldmann, J., An Optically Controlled Microscale Elevator Using Plasmonic Janus Particles. ACS Photonics 2015, 2 (4), 491-496.

17. Li, J.; Esteban-Fernández de Ávila, B.; Gao, W.; Zhang, L.; Wang, J., Micro/ nanorobots for biomedicine: Delivery, surgery, sensing, and detoxification. Science Robotics 2017, 2 (4), eaam6431.

18. Fan, X.; White, I. M., Optofluidic microsystems for chemical and biological analysis. Nature Photonics 2011, 5 (10), 591-597.

19. Weinert, F. M.; Braun, D., Optically driven fluid flow along arbitrary microscale patterns using thermoviscous expansion. J. Appl. Phys. 2008, 104 (10).

20. Weinert, F. M.; Kraus, J. A.; Franosch, T.; Braun, D., Microscale fluid flow induced by thermoviscous expansion along a traveling wave. Phys. Rev. Lett. 2008, 100 (16), 164501.

21. Mittasch, M.; Gross, P.; Nestler, M.; Fritsch, A. W.; Iserman, C.; Kar, M.; Munder, M.; Voigt, A.; Alberti, S.; Grill, S. W.; Kreysing, M., Non-invasive perturbations of intracellular flow reveal physical principles of cell organization. Nat. Cell Biol. 2018, 20 (3), 344-351.

22. Erben, E.; Seelbinder, B.; Stoev, I. D.; Klykov, S.; Maghelli, N.; Kreysing, M., Feedback-based positioning and diffusion suppression of particles via optical control of thermoviscous flows. Opt. Express 2021, 29 (19), 30272-30283.

23. Stoev, I. D.; Seelbinder, B.; Erben, E.; Maghelli, N.; Kreysing, M., Highly sensitive force measurements in an optically generated, harmonic hydrodynamic trap. eLight 2021, 1 (1), 7.

24. Liao, W.; Erben, E.; Kreysing, M.; Lauga, E., Theoretical model of confined thermoviscous flows for artificial cytoplasmic streaming. Phys. Rev. Fluids 2023, 8 (3), 034202.

25. Erben, E.; Liao, W.; Minopoli, A.; Maghelli, N.; Lauga, E.; Kreysing, M., Opto-fluidically multiplexed assembly and micro-robotics. Light Sci. Appl. 2024, 13 (1), 59.

26. Erben, E.; Saraev, I.; Liao, W.; Nan, F.; Lauga, E.; Kreysing, M., Optical Micromanipulations Based on Model Predictive Control of Thermoviscous Flows. Small 2025, e01039

27. MacLeod, B. P.; Parlane, F. G. L.; Morrissey, T. D.; Häse, F.; Roch, L. M.; Dettelbach, K. E.; Yunker, L. P. E.; Hein, J. E.; Berlinguette, C. P., Self-driving laboratory for accelerated discovery of thin-film materials. _Sci. Adv._ 2020, _6_ (20), eaaz8867.

28. Burger, B.; Maffettone, P. M.; Gusev, V. V.; Aitchison, C. M.; Bai, Y.; Wang, X.; Li, X.; Alston, B. M.; Li, B.; Clowes, R.; Rankin, N.; Harris, B.; Sprick, R. S.; Cooper, A. I., A mobile robotic chemist. _Nature_ 2020, _583_, 237-241.

29. Granda, J. M.; Donina, L.; Dragone, V.; Long, D.-L.; Cronin, L., Controlling an organic synthesis robot with machine learning to search for new reactivity. _Nature_ 2018, _559_, 377-381.

30. Kalhor, P.; Jung, N.; Bräse, S.; Wöll, C.; Tsotsalas, M.; Friederich, P., Functional Material Systems Enabled by Automated Data Extraction and Machine Learning. _Adv. Funct. Mater._ 2023, _34_ (20), 2302630.

31. Coley, C. W.; Thomas, D. A.; Lummiss, J. A. M.; Jaworski, J. N.; Breen, C. P.; et al., A robotic platform for flow synthesis of organic molecules informed by AI planning. _Science_ 2019, _365_ (6453), 557-561.

32. Boiko, D.; Chan, B.; Lu, C.; Jensen, K. F.; Aspuru-Guzik, A., Autonomous chemical research with large language models. Nature 2023, 624, 570–578

33. Nägele, M.; Marquardt, F., Agentic exploration of physics models. arXiv preprint arXiv:2509.24978 (2025).

34. Ma, Yecheng Jason; Liang, William; Wang, Guanzhi; Huang, De-An; Bastani, Osbert; Jayaraman, Dinesh; Zhu, Yuke; Fan, Linxi ("Jim"); Anandkumar, Anima. Eureka: Human-Level Reward Design via Coding Large Language Models. In: Proceedings of the 12th International Conference on Learning Representations (ICLR 2024)

35. Ma, Yecheng Jason; Liang, William; Wang, Hung-Ju; Wang, Sam; Zhu, Yuke; Fan, Linxi; Bastani, Osbert; Jayaraman, Dinesh. DrEureka: Language Model Guided Sim-To-Real Transfer. In: Proceedings of the Robotics: Science and Systems (RSS) Conference 2024

36. Paul, D.; Milosevic, N.; Scherf, N.; Cichos, F., Physical embodiment enables information processing beyond explicit sensing in active matter. arXiv preprint arXiv:2508.17921 (2025).

37. de Bakker, V.; Hejna, J.; Lum, T. G. W.; Celik, O.; Taranovic, A.; Blessing, D.; Neumann, G.; Bohg, J.; Sadigh, D., Scaffolding Dexterous Manipulation with Vision-Language Models. arXiv preprint arXiv:2506.19212 (2025).

38. Roch, L. M.; Häse, F.; Kreisbeck, C.; Tamayo-Mendoza, T.; Yunker, L. P. E.; Hein, J. E.; Aspuru-Guzik, A., ChemOS: An orchestration software to democratize autonomous discovery. _Chem. Sci._ 2018, _9_, 6091-6098.

39. Liao, Weida. Flows inside Cells: From Natural Cytoplasmic Streaming to Microfluidic Systems. Apollo - University of Cambridge Repository, 2025, doi:10.17863/CAM.122076.

40. Ghareeb, A. E.; Chang, B.; Mitchener, L.; Yiu, A.; Szostkiewicz, C. J.; Shved, D.; Gyimesi, G. J.; Laurent, J. M.; Wright, S. M.; Razzak, M. T.; White, A. D.; Finnemann, S. C.; Hinks, M. M.; Rodriques, S. G. A multi-agent system for automating scientific discovery. Nature 2026, doi:10.1038/s41586-026-10652-y.

41. Aygün, E.; Belyaeva, A.; Comanici, G.; Coram, M.; Cui, H.; Garrison, J.; Johnston, R.; Kast, A.; McLean, C. Y.; Norgaard, P.; Shamsi, Z.; Smalling, D.; Thompson, J.; Venugopalan, S.; Williams, B. P.; et al. An AI system to help scientists write expert-level empirical software. Nature 2026, doi:10.1038/s41586-026-10658-6.

42. Gottweis, J.; Weng, W.-H.; Daryin, A.; Tu, T.; Sirkovic, P.; Myaskovsky, A.; Glowaty, G.; Weissenberger, F.; Orlandi, A.; Tomašev, N.; Zverinski, D.; Rendulic, I.; Vedadi, E.; Hasler, F.; Rimanic, L.; et al. Accelerating scientific discovery with Co-Scientist. Nature 2026, doi:10.1038/s41586-026-10644-y.